\documentclass[11pt,a4paper]{article}
\usepackage{swisstextkonvens2020}
\usepackage{hyperref}
\usepackage{times}
\usepackage{latexsym}
\usepackage{comment}

\usepackage{lipsum}
\usepackage{graphicx}
\usepackage[utf8]{inputenc}
\usepackage{mathptmx}
\usepackage{hyperref}
\usepackage[scaled=.90]{helvet}
\usepackage{courier}
\usepackage{color, colortbl}
\usepackage{booktabs}
\usepackage[T1]{fontenc}
\usepackage{makecell}
\usepackage{multirow}
\usepackage{subcaption}
\usepackage{tikz}
\usepackage{pgfplots}
\pgfplotsset{compat = 1.9}
\pgfplotsset{major grid style={dashed}}
\usepackage{siunitx, mhchem}
\usepackage{amsmath}
\usepackage{url}

\newcommand{\DEVELOPMENT}{1} 
\usepackage{ifthen}
\ifthenelse{\DEVELOPMENT = 1}{
	\newcommand{\tz}[1]{\textcolor{purple}{\textbf{TZ:} #1}}	
    \newcommand{\aga}[1]{\textcolor{blue}{\textbf{AA:} #1}}
    \newcommand{\ah}[1]{\textcolor{cyan}{\textbf{AH:} #1}}
}{
	\newcommand{\tz}[1]{}
	\newcommand{\aga}[1]{}
	\newcommand{\ah}[1]{}
}

\makeatletter
\def\url@leostyle{%
  \@ifundefined{selectfont}{\def\UrlFont{\sf}}{\def\UrlFont{\scriptsize\sffamily}}}
\makeatother
\aclfinalcopy 

\title{Robustness of end-to-end Automatic Speech Recognition Models --\\ A Case Study using Mozilla DeepSpeech}

\author{Aashish Agarwal \and Torsten Zesch\\
  Language Technology Lab\\
  University of Duisburg-Essen\\
  Duisburg, Germany
  }

\date{}

\begin{document}
\maketitle
\begin{abstract}
When evaluating the performance of automatic speech recognition models, usually word error rate within a certain dataset is used.
Special care must be taken in understanding the dataset in order to report realistic performance numbers.
We argue that many performance numbers reported probably underestimate the expected error rate.
We conduct experiments controlling for selection bias, gender as well as overlap (between training and test data) in content, voices, and recording conditions.
We find that content overlap has the biggest impact, but other factors like gender also play a role.
\end{abstract}

\section{Introduction}

Automatic Speech Recognition (ASR) has made striking progress in recent years with the deployment of increasingly large deep neural networks \cite{asr_nn_6,asr_nn_2,asr_nn_5,asr_nn_7}.
Now when you see a shiny new model with an error rate reported to be below 10\%, are you likely to get the same error rate on your data?
We argue that many reported results probably underestimate the word error rate (WER) to be expected when a model is applied outside of its exact training conditions.

For example, in many datasets, there is a large imbalance between male and female voices (usually not enough female data).
When evaluating only within such a dataset and not controlling for gender, the model can optimize overall WER by performing worse for females \cite{genderbias}.
If the model is eventually applied in a setting where males and females are equally likely to use the system, WER will be much higher.

Other issues that might lead to underestimating error rate are overlaps between the train and test sets regarding content, voices or recording conditions.
Another issue to be considered is selection bias when the training process can select samples for training and testing.

A really robust model should generalize beyond these factors, but we find that current models trained on the available datasets do not.
We argue that this is partly due to the focus on reporting improvements in a within-dataset setting.
It just sounds better to report a 4.3\% WER on the standard dataset instead of a more realistic number (which we show can be several times higher).
However, as most real-world applications are unlikely to directly reflect the properties of a specific dataset, most users would be better off with more robust models and a realistic estimate.

Most of the end-to-end speech recognition systems for English use the Librispeech \cite{librispeech} corpus, which has pre-defined data splits trying to avoid the issues discussed above.\footnote{However, note that over time fixed data splits lead to overfitting the methods on the dataset.}
For German data, standard splits are not fully established leading to large differences in WER between datasets, e.g.\ \newcite{GermanDeepSpeech} report WER in the range between 15 and 79.

We argue that this is also a challenge for other languages, where standard data splits are not defined, including Arabic \cite{arabic}, Kazak \cite{Kazakh}, Bengali \cite{bengali}, and Russian \cite{russian}.

We thus perform experiments investigating the relative impact of dataset properties in order to give practical advice on how to train the models.
This might also have consequences for the way speech datasets are collected.
For data-rich languages like English, these issues can somewhat be offset by using more training data, so that a model might still be able to generalize well across different conditions.
We thus perform our experiments on German, which --at least when it comes to the amount of publicly available, transcribed speech data-- has to be counted as an under-resourced language.
We perform our experiments using the end-to-end speech recognition toolkit Mozilla DeepSpeech.\footnote{\url{https://github.com/mozilla/DeepSpeech}}
Our results probably generalize to other neural architecture similar to DeepSpeech.

We make our experimental setup publicly available (URL removed for review).

\section{Dataset Properties}

\begin{table*}[t]
    \centering
    \small
    \begin{tabular}{llcrrr}
    \toprule
                      &            &       \multicolumn{2}{c}{\bf Number of}     &                                          \multicolumn{2}{c}{\bf [h]} \\
        \cmidrule(r){3-4} \cmidrule(l){5-6}
          \bf Dataset & \bf Domain & \bf Mics                                                           & \bf Voices  & \bf Total   & \bf Unique   \\
         \midrule
         \addlinespace[2mm]
         TUDA-De (v2)  &  Wikipedia, Europarl, Commands &   5                               & 179                 & 184         & 7           \\
         \addlinespace[2mm]
        Mozilla Common Voice (MCV) v3      &  Wikipedia    & many                                                  & 4850               & 321         & 24          \\
         \addlinespace[2mm]
         M-AILABS     &  \makecell[l]{Audiobooks (LibriVox, Project Gutenberg),\\ Speeches, Interviews}     & ? &  \textasciitilde 5       & 233         & 233         \\
         \bottomrule
    \end{tabular}
    \caption{German datasets used in this study}
    \label{tab:datasets}
\end{table*}

As we argue that dataset properties play such a big role, we will first have a look at the available training data collections.
While for English or Chinese quite large datasets are publicly available, all German datasets are of limited size (see Table~\ref{tab:datasets}).

However, only focusing on the overall size is misleading anyway as e.g.\ even one million hours of one person reading the same sentence over and over again would not result in a usable model.
We thus also look at other properties.
A dataset like M-AILABS with very few voices is unlikely to generalize well to new voices.
One the other hand, a dataset like Mozilla Common Voice (MCV) with thousands of voices easily reaches the largest overall size in our set, but as most voices repeat the same sentences, the dataset does not capture the same breadth of lexical material.
As a consequence, the size of unique content in the MCV dataset is rather small, but not as small as the TUDA-De dataset where each sample is recorded by 5 different microphones bringing the unique size down to 7 hours (from 184 hours in total).

We thus argue that the question \textit{Can I train a robust model with [XYZ] hours of data?} cannot be answered without estimating the relative influence that each of these factors is going to have on the training process.

\subsection{Voice Gender}

\begin{table}[t]
    \centering
    \small
    \begin{tabular}{lrrrrrrr}
    \toprule
          & \multicolumn{2}{c}{\bf TUDA-De} & \multicolumn{2}{c}{\bf MCV} & \multicolumn{2}{c}{\bf  M-AILABS}  \\
         \cmidrule{2-7}
         \bf Gender    
         & \bf \# & \bf [h] & \bf \# & \bf [h] & \bf \# & \bf [h] \\
         \midrule
         Male          
         & 129           & 123 & 1555   & 215 & 1  &  40     \\
         Female        
         &  50           &  61 &  173   &  33 & 4  & 147    \\
         Unknown       
         &   -           &   - & 3122   &  73 & ?  &  46  \\
         \midrule
         male:female
                       & 3:1     & 2:1 & 9:1   & 7:1 & 1:4 & 1:4    \\
         \bottomrule
    \end{tabular}
    \caption{Dataset analysis regarding gender of voices}
    \label{tab:genderDistribution}
\end{table}

As we are not aware that the gender balance of the available German datasets has been analyzed in detail before, we provide the statistics in Table~\ref{tab:genderDistribution}. 
We found that across almost all the datasets, except M-Ailabs, the number of male voices is predominantly high.
For example, in TUDA-De, male to female ratio is 3:1 and in MCV it is 9:1.
This means that male voices form the majority of the corpora.  
Thus such corpora might not be able to generalise well in realistic settings.
Projects collecting speech samples from volunteers should try to recruit more women and in general a more diverse set of dialects etc.
When designing a speech corpus, keeping diversity (not only regarding gender) in mind would be beneficial.

\subsection{Data Splits}
\label{sec:splits}

\begin{figure}[t]
  \centering
  \includegraphics[scale=0.26]{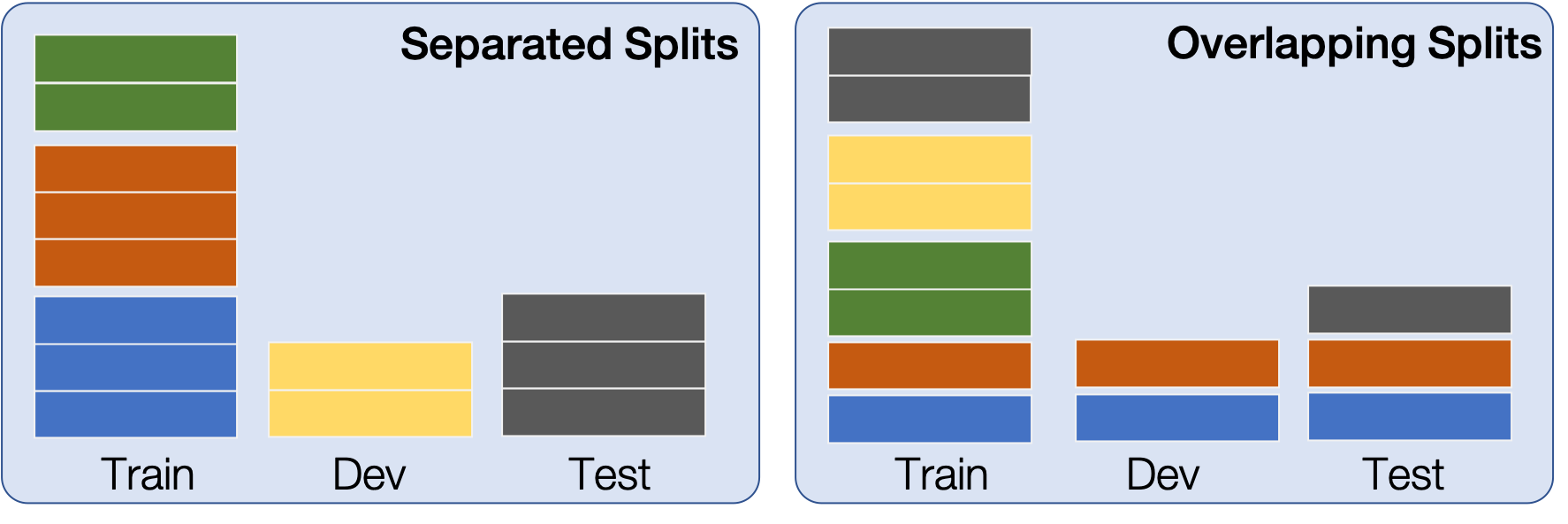}
  \caption{Visualization of data split issue}
  \label{fig:visualization_split}
\end{figure}

Having a dataset with multiple voices, varied recording conditions, and little content redundancy does not automatically guarantee a robust model.
Care has to be taken to separate cases between train, validation and test.
Figure~\ref{fig:visualization_split} visualizes the issue in a general way.
A fixed data split (left) should separate dimensions are much as possible, e.g.\ not have the same voices or the same content in train and test (right).

Of course, the severity of the issue depends on the usage scenario.
If all one wants to do is recognizing spoken digits from 0 to 10, there is no harm with having samples of all digits in train and in the test, as in the application scenario those digits are all to care about.
However, if the goal is a robust, domain-independent model, we need to control for overlap in sentences between train and test in order to obtain a realistic error rate estimate.

\subsection{Selection Bias}

\begin{figure}[t]
  \centering
  \includegraphics[scale=0.65]{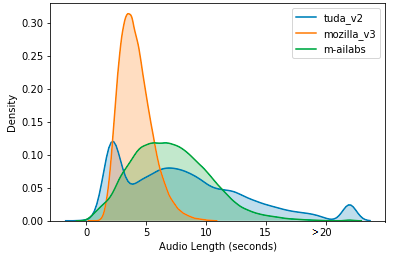}
  \caption{Distribution of sample length}
  \label{fig:length_distribution}
\end{figure}

An issue indirectly related to dataset properties is that frameworks often perform some kind of preprocessing and might filter out some samples in the process.
For example, in Figure~\ref{fig:length_distribution} we show the length distribution of samples in each dataset.
Without looking at other dataset properties it might look useful to get rid of very short or very long samples and to only train (and test!) a model using samples close to the peak of the distribution.
However, this might introduce a selection bias, where we reduce WER by simply discarding all the hard cases.
This leads to excellent within-dataset results, but poor cross-dataset results.

\section{Experiments \& Results}

For our experiments, we used the latest released version of Mozilla DeepSpeech (v0.6.0).\footnote{\url{https://github.com/mozilla/DeepSpeech/releases/tag/v0.6.0}}
We choose the best hyperparameters\footnote{Batch Size - 24, Dropout - 0.25, Learning Rate - 0.0001} as described in \cite{GermanDeepSpeech}.
The models are trained and tested on a compute server having 56 Intel(R) Xeon(R) Gold 5120 CPUs @ 2.20GHz, 3 Nvidia Quadro RTX 6000 with 24GB of RAM each.
The typical training time on a single dataset under this setup was in the range of 2 hours.
We ran our experiments for approximately 200 hours, which is equivalent to about 50 kg of CO$_2.$\footnote{\url{https://www.rensmart.com/Calculators/KWH-to-CO2}}

\subsection{Baseline: All data, random split}


As a baseline, we simply take all data and randomly split the data into train/dev/test, i.e.\ we do not take any of the dataset properties discussed above into account.
This is the setup that is most likely used whenever not discussed differently in a paper.
Table~\ref{tab:crossDomainResults} gives an overview of the WER obtained in that way (rows in italics).
Given the limited amount of training data, the results are in the expected range and generally similar to previously reported results \cite{GermanDeepSpeech}.
However, as noted above, those numbers are probably underestimating the true error rate.

\begin{table}[t]
    \centering
    \small
    \begin{tabular}{llc}
    \toprule
         \bf Train   &  \bf Test &  \bf WER \\
         \midrule
        
       \multirow{4}{*}{TUDA-De}    & \it TUDA-De (v2)  & \it 14.9 \\
                                        &   MCV (v3)        &  79.3    \\
                                        &   M-AILABS        &  79.7   \\
        \addlinespace[2mm]
         
        \multirow{4}{*}{MCV}      & \it MCV (v3)     & \it 26.8  \\
                                        &  TUDA-De (v2)    &  54.6   \\
                                        &  M-AILABS        &  43.7   \\
        \addlinespace[2mm]
        
         \multirow{4}{*}{M-AILABS}      & \it M-AILABS     & \it 17.5 \\
                                        &  TUDA-De (v2)    & 84.9     \\
                                        &  MCV (v3)        & 68.3  \\
         \bottomrule
    \end{tabular}
    \caption{Cross-domain results}
    \label{tab:crossDomainResults}
\end{table}

We thus also conduct cross-domain experiments, as testing on a dataset different from training is a natural way of checking the model robustness without any overlap at all.
If the WER reported on the dataset itself is a realistic measure of performance, we should see cross-domain results that are similar.
However Table~\ref{tab:crossDomainResults} shows that WER always dramatically rises -- mostly to the point that the model is not being useful anymore.
MCV seems to generalize somewhat better than TUDA-De or M-AILABS, which indicates that many voices are more important for model robustness than more unique training samples.

We the remainder of this section, we explore which other factors are influencing results the most.

\subsection{Content overlap}

\begin{table}[t]
    \centering
    \small
    \begin{tabular}{lrcc}
    \toprule
          \bf Dataset      & \bf [h] & \bf Baseline       & \bf No content  \\
         \midrule
              TUDA-De      & 184 & 14.9     & 66.9           \\
              MCV          & 321 & 26.8     & 43.9           \\
              M-AILABS     & 233 & 17.5     & 17.1           \\
         \bottomrule
    \end{tabular}
    \caption{WER without content overlap}
    \label{tab:nocontentoverlap}
\end{table}

\begin{table*}[t]
    \centering
    \small
    \begin{tabular}{lrrcccc}
    \toprule
     &  &   \multicolumn{2}{c}{\bf Number of Voices}    &  \multicolumn{3}{c}{\bf WER} \\
     \cmidrule(r){3-4} \cmidrule(l){5-7}
     \bf Dataset      & \bf Total Size [h] &\bf Train  & \bf Dev, Test (each)  & \bf No Content  & \bf No Voice & \bf No Content \& Voice  \\
          \midrule
              TUDA-De      &  184 &   145  &  15  & 66.9 & 37.2   & 74.1  \\
              M-AILABS     &  186 &     3  &   1  & 17.8 & 72.1   & 75.2  \\
         \bottomrule
    \end{tabular}
    \caption{Results with No Voice and No Sentence Overlap}
    \label{tab:sentenceandspeakeroverlap}
\end{table*}

Table~\ref{tab:nocontentoverlap} compares the baseline results with the setup when there is no content overlap (i.e.\ exact same utterance) between the data splits.
Note that we use the same amount of data in both conditions, only the splits are different.

M-AILABS is not affected, as there is no content overlap to begin with.\footnote{The small difference is due to the independent randomization when re-running an experiment.}
This nicely shows that the results obtained for a specific dataset are replicable in general.
The other datasets are heavily effected showing that content overlap is the main reason for underestimating the true error rate.
As the MCV dataset has many voices and microphones, the 43.9 WER is probably already a robust estimate (cf.\  cross-domain results in Table~\ref{tab:crossDomainResults}).

\subsection{Voice overlap}

Table~\ref{tab:sentenceandspeakeroverlap} first shows the results without content overlap (these are the same numbers as in Table~\ref{tab:nocontentoverlap}) and then the results without voice overlap.
The WER on M-AILABS, that only has very few voices, goes up to over 70\% well into the unusable range.
Results for TUDA-De go down, but only as we are not controlling for content overlap anymore.
This is another piece of evidence that content is actually more important than voices, as it has a relatively larger impact.
If we control for both (last column), all models perform approximately on the same abysmal level.

\subsection{Recording conditions}


TUDA-De is the only dataset where we can easily control recording conditions in the form of microphones used.\footnote{Actually `recording conditions' is a much wider variable, but not present as meta-data in most datasets.}
We can use 88h for this experiment and use 3 mics for training and 1 for dev and test each.
Without content overlap, we obtain a WER of 73.8, while without mic overlap it is 53.1.
Content overlap is thus the much more important factor.
Consequently removing content and mic overlap only slightly increases WER to 77.4.

\subsection{Gender}


As we have shown, the influence of content overlap is rather strong and likely to overshadow any gender effect to be found in the data.
We thus isolate the gender variable by creating a sub-corpus where there is not content overlap between train and test and where the test set for male and female voices contains the same sentences.
We find that training on male yields 63.5 WER for males and 87.4 for females showing the expected gender gap.
If the train only on female voices, we get 55.2 WER for females and 88.3 for males.

\section{Summary}
\label{sec:summary}
Our study shows that the robustness of end-to-end speech recognition models heavily depends on dataset splits.
Content overlap is the main reason for underestimating the true error rate.
Especially in datasets that are collected in a crowd-sourced fashion, where many voices read the same sentences, or when multiple microphones are used, extra care has to be taken to avoid information leakage from train to test.
However, other factors like gender balance or recording conditions are also contributing to the effect.

\bibliographystyle{acl_natbib}
\bibliography{asr}

\end{document}